\documentclass[letterpaper]{article} 
\usepackage{aaai2026}  
\usepackage{times}  
\usepackage{helvet}  
\usepackage{courier}  
\usepackage[hyphens]{url}  
\usepackage{graphicx} 
\usepackage{natbib}  
\usepackage{caption} 
\usepackage{algorithm}
\usepackage{algorithmic}
\usepackage{mathtools}
\usepackage{amsmath}
\usepackage{amssymb}
\usepackage{multirow}
\usepackage{makecell}
\usepackage{booktabs}
\usepackage{array}
\usepackage{tabularx}
\usepackage{cleveref}
\usepackage{adjustbox}
\newcolumntype{C}{>{\centering\arraybackslash}X}
\usepackage{dblfloatfix}
\usepackage{newfloat}
\usepackage{listings}
\DeclareCaptionStyle{ruled}{labelfont=normalfont,labelsep=colon,strut=off} 
\floatstyle{ruled}
\newfloat{listing}{tb}{lst}{}
\floatname{listing}{Listing}
\title{HCC-3D: Hierarchical Compensatory Compression for 98\% 3D Token Reduction in Vision-Language Models}
\author{
    Liheng Zhang\textsuperscript{\rm 1},
    Jin Wang\textsuperscript{\rm 1},
    Hui Li\textsuperscript{\rm 2},
    Bingfeng Zhang\textsuperscript{\rm 1}\thanks{Corresponding author.},
    Weifeng Liu\textsuperscript{\rm 1}
}
\affiliations{
    \textsuperscript{\rm 1}China University of Petroleum (East China)\\
    \textsuperscript{\rm 2}The Hong Kong Polytechnic University\\
    \{lihengzhang, wangjin\}@s.upc.edu.cn, hui5li@polyu.edu.hk, \{Bingfeng.Zhang, liuwf\}@upc.edu.cn
}

\usepackage{bibentry}

\begin{document}

\maketitle

\begin{abstract}
3D understanding has drawn significant attention recently, leveraging Vision-Language Models (VLMs) to enable multi-modal reasoning between point cloud and text data. Current 3D-VLMs directly embed the 3D point clouds into 3D tokens, following large 2D-VLMs with powerful reasoning capabilities. However, this framework has a great computational cost limiting its application, where we identify that the bottleneck lies in processing all 3D tokens in the Large Language Model (LLM) part. This raises the question: how can we reduce the computational overhead introduced by 3D tokens while preserving the integrity of their essential information? To address this question, we introduce Hierarchical Compensatory Compression (HCC-3D) to efficiently compress 3D tokens while maintaining critical detail retention. Specifically, we first propose a global structure compression (GSC), in which we design global queries to compress all 3D tokens into a few key tokens while keeping overall structural information. Then, to compensate for the information loss in GSC, we further propose an adaptive detail mining (ADM) module that selectively recompresses salient but under-attended features through complementary scoring. Extensive experiments demonstrate that HCC-3D not only achieves extreme compression ratios (approximately 98\%) compared to previous 3D-VLMs, but also achieves new state-of-the-art performance, showing the great improvements on both efficiency and performance.
\end{abstract}

\begin{links}
    \link{Code}{https://github.com/lihengzhang02/HCC-3D}
\end{links}

\section{Introduction}

Large Vision-Language Models (VLMs) have brought revolutionary changes to artificial intelligence by integrating language and visual information~\cite{openai2022chatgpt,huang2024audiogpt,cheng2024videollama}. Despite these advances, extending multimodal capabilities to three-dimensional understanding remains a core challenge. To fill this critical gap, recent studies have proposed 3D-VLMs capable of directly sensing and reasoning about 3D point cloud data~\cite{hong20233dllm,xu2024pointllm,qi2024shapellm}, thus bridging 3D perception and language understanding.

\begin{figure}[t!]
\centering
\includegraphics[width=0.48\textwidth]{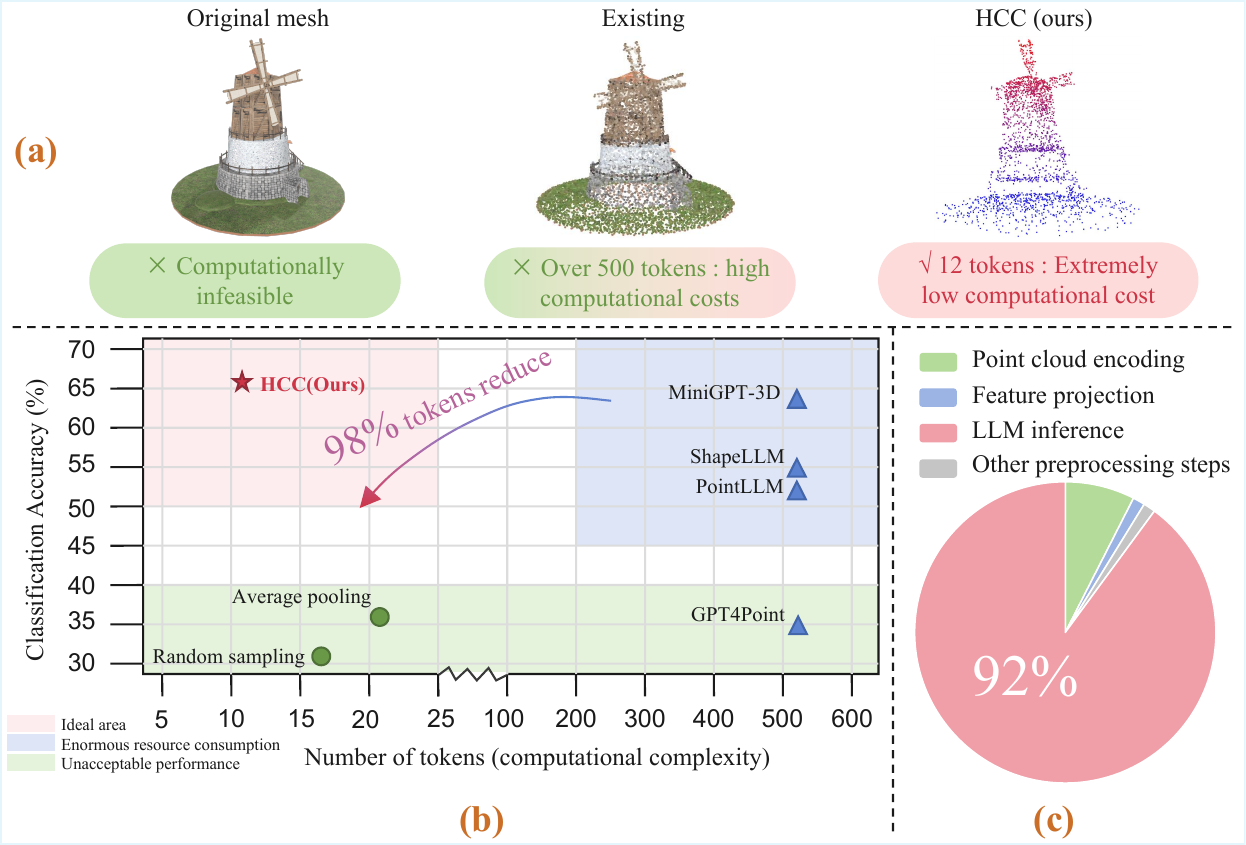}
\caption{\textbf{Performance comparison of 3D point cloud tokenization methods.} (a) 3D Token Compression: HCC achieves 12 tokens vs. 500+ in existing Methods. (b) Relationship chart between token count and classification accuracy. Our HCC-3D uses less 3D tokens yet maintains higher performance. (c) Proportion of inference time. The LLM part of the current 3D VLMs takes over 90\% computing costs. Best view in color.}
\label{fig:mo}
\end{figure}

The current mainstream approach to 3D-VLMs employs a paradigm that integrates point cloud representations into existing 2D-VLMs architectures. These methods typically employ specialized 3D encoders to extract point cloud features, which are then projected into the embedding space of pre-trained 2D VLMs through learned alignment~\cite{xu2024pointllm,tang2024minigpt3d}. For example, PointLLM~\cite{xu2024pointllm} utilizes a point cloud encoder followed by a linear projection layer to map 3D features into the input space of LLaMA~\cite{touvron2023llama}, while 3D-LLM~\cite{hong20233dllm} employs a more complex alignment network to bridge the representation gap. These methods demonstrate the feasibility of enabling VLMs to comprehend 3D data by leveraging the powerful reasoning capabilities of existing 2D-VLMs. However, while these methods have shown promising results in 3D understanding, the computational burden imposed on VLMs by processing all high-dimensional 3D visual tokens remains a key obstacle that limits their practical deployment.

To effectively address the computational challenges posed by massive 3D visual tokens, we first examine where the computational bottlenecks lie. Previous research has demonstrated that the computational overhead is predominantly concentrated in the LLM component~\cite{men2024shortgpt}. As illustrated in Fig.~\ref{fig:mo}(c), the LLM processing accounts for over 90\% of the total inference time, primarily due to processing tokens through multiple Transformer layers. Therefore, mitigating the computational burden of LLMs is crucial for enhancing the utility and scalability of 3D-VLMs. One intuitive strategy is to reduce the number of tokens input to the language model. Although token reduction has been extensively explored in 2D vision-language tasks, empirical evidence indicates that naive token compression often results in substantial performance degradation~\cite{li2024inference}. This trade-off between efficiency and accuracy is particularly pronounced in 3D scenes due to the inherent characteristics of point cloud data. This observation raises a fundamental challenge: How can we reduce the computational overhead introduced by 3D tokens while preserving the integrity of their essential information?

To satisfy the above requirement, when designing compression strategies for 3D point clouds, their unique spatial characteristics must be considered: Their irregular distribution leads to heterogeneous information density, with some regions encoding rich geometric details while others remain sparse and redundant. To maintain a holistic structural understanding and preserve task-relevant information, compression strategies should be adaptive, and they must incorporate specialized designs to keep critical data intact for downstream tasks. 

Based on this insight, we propose a Hierarchical Compensatory Compression (HCC-3D) method for 3D VLMs, which employs a dual-path architecture, \emph{i.e.}, Global Structure Compression (GSC) and Adaptive Detail Mining (ADM), to build global structure preservation with adaptive detail mining. For the GSC, we design learnable 3D spatial queries with multi-head attention mechanism to achieve overall compression while preserving the basic geometric structure of 3D objects. For the ADM, we design an attention-guided selection mechanism to dynamically identify regions that are under-attended but informationally rich, followed by a recompression operation using dedicated detail queries, to preserve critical yet overlooked information. 
As shown in Fig.~\ref{fig:mo}(a) and (b), this divide-and-conquer design enables HCC-3D to achieve extreme compression efficiency, reducing 3D features to a minimal number of tokens—a breakthrough that significantly reduces training time compared to existing methods. Moreover, our hierarchical complementarity mechanism effectively reduces information loss during compression. When global queries miss important regions, the detail queries identifies and preserves these areas, creating a representation that is both comprehensive and compact while maintaining high quality for various downstream tasks.

In summary, our contributions are as follows:

\begin{itemize}
\item We discover that the computational bottleneck of 3D-VLMs lies in LLM processing massive 3D visual tokens. To address this, we propose Hierarchical Compensatory Compression (HCC-3D), a novel method for efficient 3D feature compression that achieves extreme reduction to merely 12 tokens.

\item We design two complementary modules: Global Structure Compression that uses spatial queries to maintain geometric structures, while Adaptive Detail Mining identifies and preserves important local regions overlooked by global compression.

\item Experiments demonstrate our approach outperforms other previous approaches with clear margin across multiple 3D tasks even at 98\% compression rates.
\end{itemize}

\begin{figure*}[t]
    \centering
    \includegraphics[width=\textwidth]{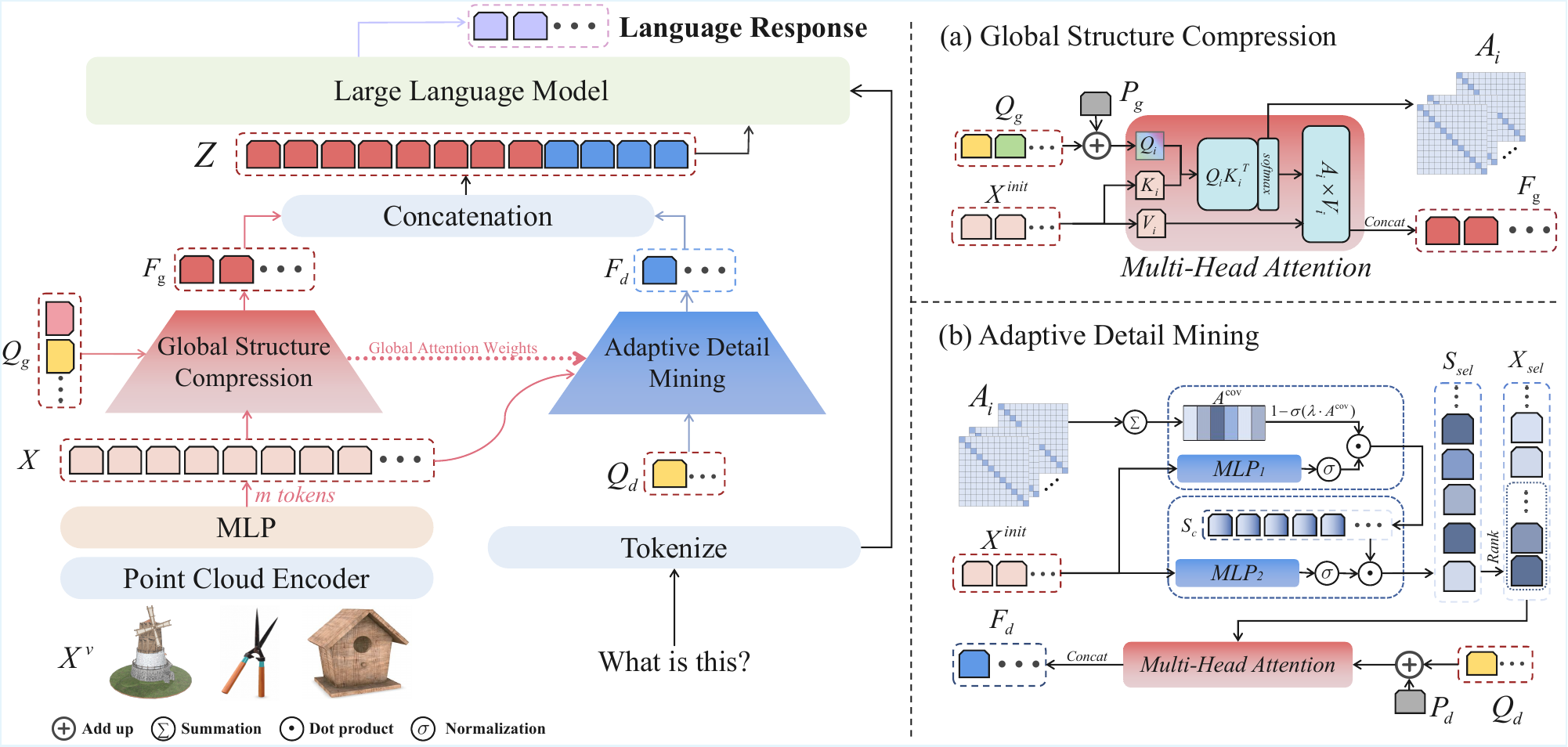}
    \caption{Overall architecture of HCC-3D. \textbf{Left}: HCC-3D compresses the 513 tokens output by the point cloud encoder into 12 tokens. \textbf{Right}: (a) Global structure (GSC) compression compress voxel features into global features and output global attention weights through a multi-head attention mechanism. (b) Adaptive Detail Mining (ADM) selects complementary features by leveraging attention weights and intrinsic feature importance.}
    \label{fig:architecture}
\end{figure*}

\section{Related Work}
\subsection{Multimodal Large Language Models}
Recent advances in large language models enabled the development of VLMs that integrate ~\cite{wang2024qwen2,li2023blip,alayrac2022flamingo}, audio~\cite{chang2023llm4ts,huang2024audiogpt}, and video~\cite{zhang2025llava,chen2023videollm} modalities for cross-modal reasoning. These models typically employed pre-trained encoders to extract features from different modalities, subsequently projecting them into the language model's embedding space through learnable projection layers to enable cross-modal understanding and generation. While these approaches achieved significant success in efficiently processing visual~\cite{mentzer2020high,chen2024generative} and audio~\cite{zeghidour2021soundstream} data, most existing frameworks still faced challenges in effectively extracting 3D visual features with comparable efficiency and scalability.

\subsection{3D Vision-Language Models}
Recent work explored integrating 3D point clouds into vision-language tasks~\cite{hong20233dllm,xu2024pointllm,qi2024shapellm}, but existing methods faced several limitations. Some approaches relied heavily on 2D representations~\cite{hong20233dllm,zhu20233d}, limiting their semantic understanding of 3D structures. Although recent work pursues direct encoding of 3D data~\cite{chen2024ll3da,zhu2023pointclip,xu2024pointllm,qi2024shapellm}, the high-dimensional and unstructured nature of point cloud features introduced substantial computational and storage overhead, limiting their application in large-scale multimodal frameworks. Our approach addresses this challenge by eliminating redundant information while preserving semantically critical regions in point cloud features, enabling effective integration with VLM architectures.

\subsection{Visual Feature Compression}
Contemporary visual feature compression research primarily focused on 2D image representations, establishing mature techniques including spatial attention mechanisms~\cite{li2024learned,zhang2021exploring}, feature pyramids~\cite{lin2017feature}, and learnable token compression~\cite{liu2022adaptive,wang2024efficient,zhang2025top}. However, point cloud compression methods remained notably insufficient, with existing approaches largely being direct adaptations of 2D methods that fail to fully consider the unique characteristics of 3D spatial data~\cite{beemelmanns20223d}. The inherent sparsity and irregularity of point clouds introduced unique challenges that traditional compression techniques cannot effectively address. This work proposes a targeted 3D compression strategy that explicitly separates global geometric structure preservation from local detail preservation, specifically addressing the computational bottlenecks inherent in point cloud processing for multimodal applications.

\section{Method}
\subsection{Overview}
To efficiently compress the point cloud features while preserving the global structural information and key local details, we propose a hierarchical compensated compression architecture. Our architecture consists of two complementary modules: a global structure compression module that uses learnable 3D spatial queries to preserve overall geometric structures, and a adaptive detail mining module that employs an attentional mechanism to dynamically identify and retain task-critical regions that be overlooked during global compression. Fig.~\ref{fig:architecture} illustrates the HCC-3D architecture with the following steps:

(1) Given an input point cloud, it is sent to a pre-trained 3D encoder to generate dense feature representations, and these representations are projected via an MLP.

(2) Then, the projected features are input to our GSC module, serving as the key and value features, while a set of learnable global queries with positional embeddings are designed as queries in the multi-head cross-attention mechanism~\cite{vaswani2017attention} to build the first compression.

(3) After that, the ADM module uses the projected features and weights from GSC to compute complementary attention scores that identify regions with local global coverage but high feature importance, then it selects the top-K features and compress them via detail queries to achieve hierarchical token reduction.

(4) Finally, the global and detail features are concatenated and fed into the language model component of the 3D-VLM for efficient multimodal comprehension and response generation.

\subsection{Global Structure Compression}

To reduce the computational overhead of processing high-dimensional 3D features, we propose a \textbf{Global Structure Compression (GSC)} module to efficiently compress point cloud features while preserving essential geometric structures. Suppose the initial point clouds are $X^v$, after passing an point-cloud encoder, \textit{m} corresponding point cloud features $X^{init} \in \mathbb{R}^{m \times d_{init}}$ are obtained. Then, $X^{init}$ is projected to $X  \in \mathbb{R}^{m \times d}$ through a projection layer $f_{MLP}$, where $d$ is the input dimension of the compression module. To capture global structural information, we design $n_g$ learnable global queries $Q_g \in \mathbb{R}^{n_g \times d}$ ($n_g \ll m$), which are enhanced with learnable positional encoding $P_g \in \mathbb{R}^{n_g \times d}$ to distinguish different global query positions.

The enriched global queries ${\hat{Q}_g} = Q_g + P_g$ are then used to attend to the entire feature sequence. To capture diverse patterns, we employ multi-head attention with $H$ heads, where each attention head is computed as:

\begin{equation}
head_i = A_i V_i = \left[ {softmax}\left( \frac{Q_i K_i^T}{\sqrt{d_k}} \right) \right] V_i
\end{equation}

where $A_i \in \mathbb{R}^{n_g \times m}$ denotes the attention weights for the $i$-th head, with $ Q_i = \hat{Q}_g W_i^Q $, $ K_i = X W_i^K $, and $ V_i = X W_i^V $ representing the query, key, and value respectively; $ W_i^Q \in \mathbb{R}^{d \times d_k} $, $ W_i^K \in \mathbb{R}^{d \times d_k} $, and $ W_i^V \in \mathbb{R}^{d \times d_v} $ are learned projection matrices, and $ d_k = d_v = d/H $.

The final global features are obtained by concatenating all heads, followed by an output projection:
\begin{equation}
F_g = \text{Concat}({head}_1, ..., {head}_{H}), \label{eq: F_g}
\end{equation}

where $\text{Concat}(\cdot,\cdot)$ denotes concatenation along the token dimension and $F_g \in \mathbb{R}^{n_g \times d}$ represents the compressed global features. This operation yields $n_g$ features that encode the overall shape structure.

\subsection{Adaptive Detail Mining}

While global compression captures coarse structural information, fine-grained details may be under-represented due to limited query token numbers. To address this, we propose an \textbf{Adaptive Detail Mining} (ADM) mechanism that identifies and recompress informative local features missed in GSC. This section contains an attention gate that identifies under-attended regions through coverage and importance scores, while the dynamic feature selection and compression selects top-K informative features based on complementary scores and compresses them using learnable detail queries.

\paragraph{\textit{Attention Gate}}
We first analyze the overall attention coverage of each input token across all attention heads and global queries. For each head $i$, we have $A_i \in \mathbb{R}^{n_g \times m}$, where $n_g$ is the number of compressed global features and $m$ is the number of original features. We compute the total coverage of each input token by aggregating across all heads and global queries:
\begin{equation}
A^{{cov}} = \sum_{i=1}^{H} \sum_{j=1}^{n_g} A_{i}^{j,:} \in \mathbb{R}^{m}
\end{equation}
where $H$ is the number of attention heads and $A_{i}^{j,:}$ denotes the $j$-th row of the $i$-th attention head.

Next, we assess the intrinsic importance of each feature using a learnable multi-layer perception $\text{MLP}_1$ for scoring:
\begin{equation}
I = \sigma\left(\text{MLP}_1(X)\right) \in \mathbb{R}^{m},
\end{equation}
where $\sigma(\cdot)$ denotes the sigmoid function.

To identify regions with low global attention but high intrinsic importance, we define the complementary score:
\begin{equation}
S_c = I \odot \left(1 - \sigma(\lambda \cdot A^{cov})\right) \in \mathbb{R}^{m},
\end{equation}
where $\odot$ represents element-wise multiplication and $\lambda$ is a scaling factor that controls the sharpness of the coverage mapping. $S_c$ emphasizes features that are semantically significant yet insufficiently attended by global queries.

\paragraph{\textit{Dynamic Feature Selection and Compression}}
We further refine the complementary scores by incorporating additional feature-specific importance:
\begin{equation}
S_{sel} = S_c \odot \sigma\left(\text{MLP}_2(X)\right) \in \mathbb{R}^{m},
\end{equation}
where $\text{MLP}_2$ is another learnable scoring network. Using these refined scores, we select the top-$K$ most informative features:
\begin{equation}
\mathcal{I} = \{t_1, t_2, \ldots, t_K\} = \text{Top}K(S_{sel}, K),
\end{equation}
\begin{equation}
X_{sel} = \{X_{t} : t \in \mathcal{I}\} \in \mathbb{R}^{K \times d},
\end{equation}

where $\mathcal{I}$ denotes the selected indices and $X_{sel}$ contains the corresponding features.

These selected features $X_{sel} \in \mathbb{R}^{K \times d}$ are then compressed using $n_d$ learnable detail queries $Q_d \in \mathbb{R}^{n_d \times d}$, which are enhanced with learnable positional encoding $P_d \in \mathbb{R}^{n_d \times d}$ to distinguish different detail query positions. The enriched detail queries ${\hat{Q}_d} = Q_d + P_d$ are used to attend to the selected features. We employ multi-head attention with $H$ heads to capture fine-grained patterns, where each attention head is computed as:
\begin{equation}
\resizebox{0.88\columnwidth}{!}{$
\begin{aligned}
{head}_i = {softmax}\Bigg( &\frac{{\hat{Q}_d} W_i^Q \cdot (X_{sel} W_i^K)^T} {\sqrt{d_{k}}} \Bigg) \cdot X_{sel} W_i^V.
\end{aligned}
$}
\end{equation}

Similar to Eq.~(\ref{eq: F_g}), the outputs of all heads are concatenated to obtain $F_d \in \mathbb{R}^{n_d \times d}$. This two-stage compression achieves a substantial reduction in token count ($m \gg K \gg n_d$) while preserving critical details.

Finally, the final compressed representation is obtained by concatenating global and detail features:
\begin{equation}
Z = \delta\left( W^{fuse} \cdot \text{Concat}(F_g, F_d) + b \right).
\end{equation}
where $\delta(\cdot)$ is the GeLU smooth activation function~\cite{hendrycks2016gaussian}, $W^{fuse} \in \mathbb{R}^{d \times d}$ is the projection weight matrix, and $b \in \mathbb{R}^{d}$ is the bias term.

\begin{table*}[!htb]
\centering
\caption{Generative 3D object classification results on the ModelNet40 test split and Objaverse. The accuracy (\%) under the Instruction-typed (I) prompt ``What is this?'' and the Completion-type (C) prompt ``This is an object of'' are reported. The \textbf{bold} and \underline{underline} indicate the best and second best results, respectively.}
\label{tab:classification}
\resizebox{\textwidth}{!}{
\begin{tabular}{lcccccccccccc}
\midrule
\multirow{2}{*}{Model} & 
\multirow{2}{*}{Pub.} & 
\multirow{2}{*}{\thead{LLM \\ Size}} & 
\multirow{2}{*}{\thead{3D token \\ count}} & 
\multicolumn{3}{c}{ModelNet40} & 
\multicolumn{3}{c}{Objaverse} & 
\multirow{2}{*}{Average} \\
\cmidrule(lr){5-7} \cmidrule(lr){8-10}
 &  &  &  &  (I) & (C) & Average & (I) & (C) & Average &  \\ 
\midrule
InstructBLIP-7B  & NeurIPS23 & 7B  & 2D & 17.67 & 22.81 & 20.24 & 21.50 & 26.00 & 23.75 & 22.00 \\
InstructBLIP-13B & NeurIPS23 & 13B  & 2D & 21.56 & 21.92 & 21.74 & 21.50 & 21.50 & 21.50 & 21.62 \\
LLaVA-1.5-7B & NeurIPS23 & 7B  & 2D & 27.11 & 21.68 & 24.40 & 37.50 & 30.00 & 33.75 & 29.07 \\
LLaVA-1.5-13B & NeurIPS23 & 13B  & 2D & 27.11 & 27.76 & 27.44 & 39.50 & 35.50 & 37.50 & 32.62 \\
GPT-4o mini & OpenAI & -  & 2D & 22.00 & 23.10 & 22.55 & 39.00 & 35.00 & 37.00 & 29.78 \\
\midrule
Point-Bind LLM & arXiv23 & 7B  & - & 46.60 & 45.02 & 45.81 & 7.50 & 7.58 & 7.54 & 26.68 \\
GPT4Point & CVPR24 & 2.7B  & 513 & 21.39 & 21.07 & 21.23 & 49.00 & 46.50 & 47.75 & 34.49 \\
PointLLM-7B & ECCV24 & 7B  & 513 & 51.34 & 50.36 & 50.85 & 62.00 & 63.00 & 62.50 & 56.68 \\
PointLLM-13B & ECCV24 & 13B  & 513 & 51.70 & 52.67 & 51.84 & 61.50 & 63.00 & 62.25 & 57.22 \\
ShapeLLM-7B & ECCV24 & 7B  & 512 & - & - & 52.15 & - & - & 62.50 & 57.33 \\
ShapeLLM-13B & ECCV24 & 13B  & 512 & - & - & 50.96 & - & - & 62.25 & 56.61 \\
MiniGPT-3D & MM24 & 2.7B  & 513 & \underline{61.99} & \underline{60.49} & \underline{61.24} & \underline{65.00} & \underline{68.50} & \underline{66.75} & \underline{64.00} \\
\midrule
\textbf{HCC-3D (Ours)}    & AAAI26        & 2.7B             & \textbf{12} & \textbf{62.72} & \textbf{61.83} & \textbf{62.28} & \textbf{67.00} & \textbf{68.50} & \textbf{67.75} & \textbf{65.02} \\
\midrule
\end{tabular}}
\end{table*}

\begin{table*}[!htb]
\caption{3D Object Captioning Results on Objaverse, the results are from Qwen2 evaluation, and traditional metrics.}
\label{tab:caption}
\resizebox{\textwidth}{!}{
\begin{tabular}{lcccccccc}
\midrule
Model     & Pub. & \thead{LLM Size}                 & \thead{3D token count} & Qwen2 & \thead{Sentence-BERT} & SimCSE  & Average \\
\midrule
InstructBLIP-7B  & NeurIPS23 & 7B            & 2D              & 16.10 & 35.79         & 36.67 & 29.52  \\
InstructBLIP-13B  & NeurIPS23 & 13B           & 2D              & 13.79 & 33.52         & 35.60 & 27.64 \\
LLaVA-1.5-7B  & NeurIPS23 & 7B            & 2D              & 17.80 & 39.32         & 41.08  & 32.73 \\
LLaVA-1.5-13B  & NeurIPS23 & 13B           & 2D              & 16.00 & 39.64         & 40.90 & 32.18 \\
GPT-4o mini   & OpenAI    & -            & 2D              & 26.00 & 38.70         & 39.13 & 34.61 \\
\midrule
Point-Bind LLM      & arXiv23 & 7B       & 512              & 1.93 & 27.29         & 25.35 & 18.19 \\
GPT-4o mini    & CVPR24 & 2.7B   &    513              & 21.75 & 41.10         & 41.24 & 34.70 \\
PointLLM-7B     & ECCV24    & 7B              & 513            & 42.20 & 48.50         & 48.92 & 46.54  \\
PointLLM-13B     & ECCV24    & 13B        & 513            & 40.40 & 49.07         & 48.41 & 45.96 \\
ShapeLLM-7B     & ECCV24    & 7B           & 512            & -     & 48.20             & 49.23  &  -    \\
ShapeLLM-13B    & ECCV24    & 13B         & 512            & -     & 48.52             & 49.98  &  -    \\
Minigpt-3D      & MM24      & 2.7B         & 513            & \underline{48.17} & \underline{49.54}         & \textbf{51.39} & \underline{49.7} \\
\midrule
\textbf{HCC-3D (Ours)}    & AAAI26         & 2.7B         & \textbf{12}             & \textbf{48.72} & \textbf{50.89}         & \underline{50.89} & \textbf{50.15} \\
\midrule
\end{tabular}}
\end{table*}

\begin{table*}[t]
\centering
\caption{Generative 3D object classification results on the ModelNet40 test split and Objaverse, note that all results are obtained using the \textbf{90k data.} following the setting in GreenPLM ~\cite{tang2025moretext}.}
\label{tab:greenplm}
\resizebox{\textwidth}{!}{
\begin{tabular}{lcccccccccccc}
\midrule
\multirow{2}{*}{Model} & 
\multirow{2}{*}{Pub.} & 
\multirow{2}{*}{\thead{LLM \\ Size}} & 
\multicolumn{3}{c}{ModelNet40} & 
\multicolumn{3}{c}{Objaverse} & 
\multirow{2}{*}{Average} \\
\cmidrule(lr){4-6} \cmidrule(lr){7-9}
 &  &  & (I) & (C) & Average & (I) & (C) & Average &  \\ 
\midrule
PointLLM-7B  & ECCV24 & 7B & 45.22 & 39.30 & 42.26 & 59.00 &  53.00 & 56.00 & 49.13 \\
MiniGPT-3D  & MM24 & 2.7B & 43.56 & 43.03 & 43.30 & 54.50 & 55.00 & 54.75 & 49.02 \\
GreenPLM  & AAAI25 & 3.8B & \underline{58.95} & \textbf{62.36} & \underline{60.66} & \textbf{60.50} &  \underline{58.50} & \underline{59.50} & \underline{60.08} \\
\midrule
\textbf{HCC-3D (GreenPLM)}    & AAAI26        & 3.8B   & \textbf{61.06} & \underline{60.37} & \textbf{60.72} & \underline{59.00} & \textbf{63.50} & \textbf{61.25} & \textbf{60.99} \\
\midrule
\end{tabular}}
\end{table*}

\section{Experiment}
\subsection{Datasets and Evaluation Metrics}
We evaluate our proposed HCC-3D method on two widely-adopted 3D understanding benchmarks: ModelNet40~\cite{wu20153d} and Objaverse~\cite{deitke2023objaverse}. ModelNet40 is a standard 3D shape classification dataset containing 12,311 clean CAD models spanning 40 object categories including furniture, vehicles, and household items. Objaverse represents a more challenging large-scale dataset with diverse real-world 3D objects, enabling comprehensive evaluation on both classification and captioning tasks.
We evaluate our model using task-specific protocols. For 3D object classification, we employ two prompt strategies: Instruction-typed (I) prompt ``What is this?'' and the Completion-type (C) prompt ``This is an object of''. Accuracy is measured via semantic matching with ground-truth labels using Qwen2 \cite{wang2024qwen2}. For 3D object captioning, we use three metrics: Qwen2-72B semantic similarity \cite{wang2024qwen2}, Sentence-BERT embedding similarity \cite{reimers2019sentence}, and SimCSE contextual alignment score \cite{gao2021simcse}.

\subsection{Implementation details}
We adopt Phi-2 ~\cite{phi2} (2.7B parameters) as the LLM backbone. Point cloud features are extracted using Point-BERT ~\cite{yu2022pointbert} pre-trained on ULIP-2 ~\cite{xue2024ulip}, producing 513 tokens of 384 dimensions. These features are projected to 2560 dimensions through a KV projection layer.
HCC-3D uses 8 learnable global queries ($n_g=8$) with sinusoidal positional embeddings, and 8-head cross-attention with LayerNorm applied to queries, keys, and values. For adaptive detail mining, an attention gate computes coverage scores using $\lambda=10$ and temperature 0.1. Importance scorer $\text{MLP}_1$ and detail selector $\text{MLP}_2$ both use Linear (2560 → 640) → GeLU → Linear (640 → 1).
The top 96 features ($K=96$) are selected and compressed into 4 detail queries ($n_d=4$) using 8-head cross-attention. The final representation combines 8 global and 4 local queries.
The output projection uses Linear (2560 → 2560) → GeLU → LayerNorm (2560). All weight matrices are Xavier-initialized with zero-initialized biases.
Notably, our full training completes in just 11.9 hours on a single RTX 4090 24GB GPU as shown in Tab. ~\ref{tab:time}, demonstrating high efficiency.

\begin{figure*}[b!]
    \centering
    \includegraphics[width=\textwidth]{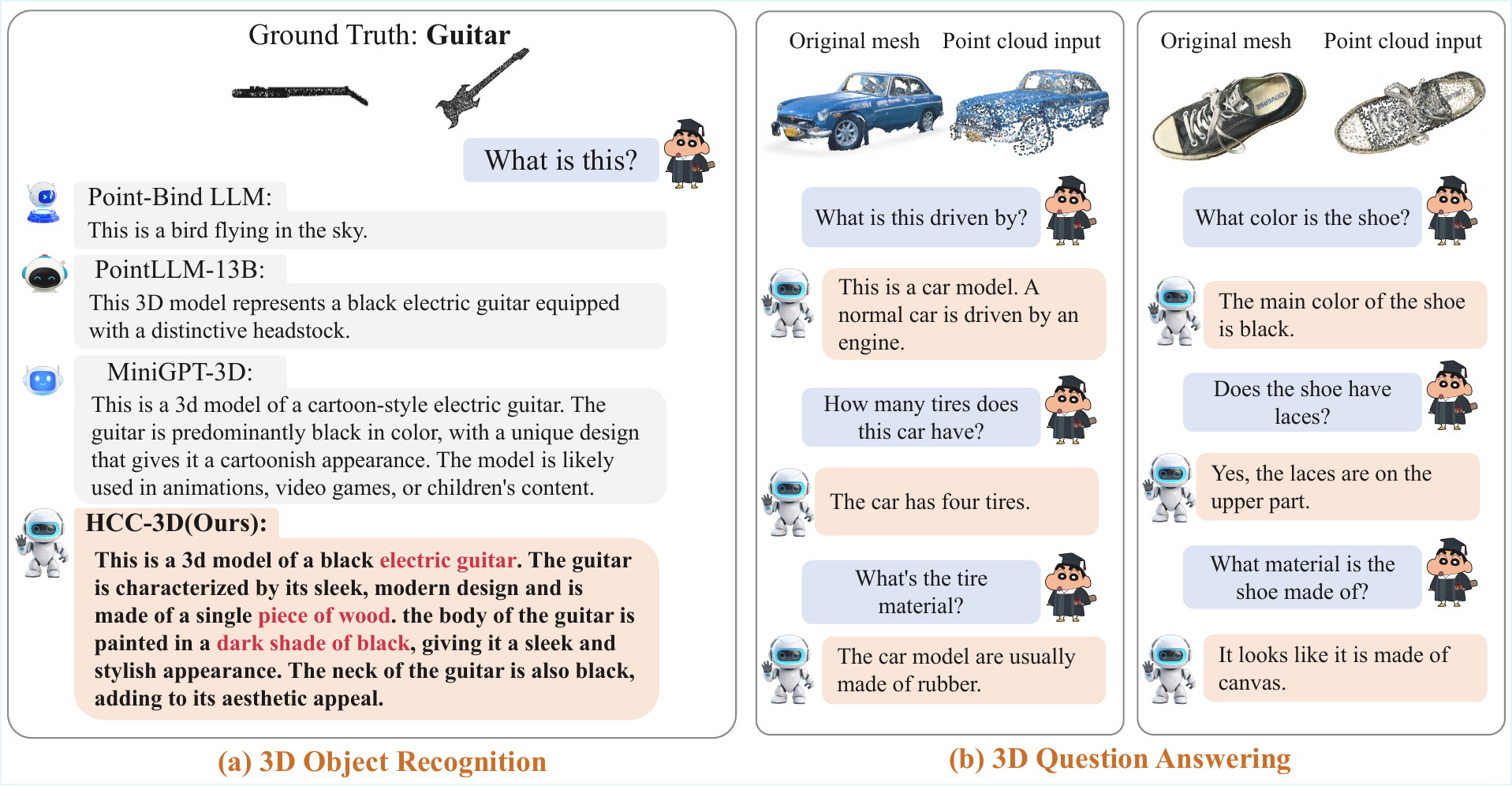}
    \caption{\textbf{Qualitative results on 3D object understanding tasks.} (a) 3D object recognition: comparison between different VLMs on identifying objects (guitar and sofa) from point cloud inputs. (b) 3D question answering: examples showing model responses to questions about 3D object properties including type, color, and material composition.}
    \label{fig:results}
\end{figure*}

\subsection{Comparison with State-of-the-art}

We compare our method with several state-of-the-art approaches for 3D understanding. For 2D-based methods, we evaluate against InstructBLIP~\cite{cui2024drive}, LLaVA-1.5~\cite{liu2023visual}, and GPT-4o mini~\cite{menick2024gpt}. For 3D-based methods, our comparison includes Point-Bind LLM~\cite{guo2023pointbind}, GPT4Point~\cite{qi2024gpt4point}, PointLLM~\cite{xu2024pointllm}, ShapeLLM~\cite{qi2024shapellm}, and MiniGPT-3D~\cite{tang2024minigpt3d}. Additionally, we integrate our approach with GreenPLM~\cite{tang2025moretext} to demonstrate its effectiveness in few-shot learning scenarios.

\paragraph{Quantitative Results.} Tab.~\ref{tab:classification} presents comprehensive comparisons on ModelNet40 and Objaverse classification tasks, note that all results are obtained using the full dataset. Our HCC-3D establishes new state-of-the-art performance while utilizing 97.6\% fewer visual tokens than other methods. On ModelNet40, HCC-3D achieves 62.28\% average accuracy across both prompt types, surpassing MiniGPT-3D by 1.04\% absolute improvement. The performance gap widens on the more challenging Objaverse dataset, where HCC-3D outperforms MiniGPT-3D by 1.00\%, achieving an accuracy of 67.75\%. Remarkably, these gains are achieved with 52\% faster training time and significantly reduced number of tokens.

Tab. ~\ref{tab:caption} summarizes the 3D captioning results on Objaverse. 
Under extreme token compression, HCC-3D achieves competitive performance, with a Qwen2 score of 48.72, which is 0.55 higher than MiniGPT-3D, and a Sentence-BERT similarity of 50.89, reflecting a 1.35 improvement. Despite a slight decrease of 0.55 in SimCSE relative to MiniGPT-3D, the overall improvements in other metrics and enhanced efficiency demonstrate the effectiveness of the proposed compression strategy.

For make fair comparison, we integrate our HCC module into GreenPLM~\cite{tang2025moretext} framework using Phi-3~\cite{abdin2024phi3} (3.8B) as the LLM backbone following GreenPLM~\cite{tang2025moretext}, which is specifically designed for efficient point cloud understanding with limited training data, note that all results are obtained using the 90k training data. Tab. ~\ref{tab:greenplm} shows that the integration achieves 60.72\% accuracy on ModelNet40 and 61.25\% on Objaverse under few-shot settings, demonstrating that our compression strategy not only generalizes across different architectures but also maintains strong performance even when trained with minimal 3D data.

\begin{table}[h]
\centering
\caption{Comparison of Training Time and Inference Speed among Different 3D-VLM Methods. Inference speed is measured as the average time per task completion.}
\label{tab:time}
\begin{tabularx}{\linewidth}{ccCC}
\toprule
Method                     & GPU & Training Time & Inference Speed \\
\midrule
PointLLM-13B            & 8*A100 (80G)          &  213h  &     $\sim 3.45\text{s}$       \\
ShapeLLM-13B   & 8*A800 (80G)          &   160h &     $\sim 2.04\text{s}$        \\
MiniGPT-3D           & 1*4090 (24G) &   16.8h &     0.45s       \\
HCC (Ours)   & 1*4090 (24G) &   \textbf{11.9h} &     \textbf{0.36s}       \\
\bottomrule
\end{tabularx}
\end{table}

\paragraph{Qualitative Analysis.}
Fig.~\ref{fig:results} demonstrates that HCC-3D achieves superior 3D understanding capabilities compared to existing methods, while InstructBLIP misidentifies the guitar as a telescope and Point-Bind LLM hallucinates entirely, HCC-3D accurately recognizes both the object category and fine-grained attributes. Notably, HCC-3D captures comprehensive details including material, design characteristics, and component-level features, this precision is achieved through our hierarchical architecture, where global compression captures overall geometry while adaptive detail mining preserves critical local features overlooked by global queries, all within just 12 tokens.
\begin{table}[h]
\centering
\caption{Ablation study on GSC and ADM modules.}
\label{tab:hcc}
\begin{tabularx}{\linewidth}{CCCC}
\toprule
GSC                     & ADM & Total & Avg. Acc.   \\
\midrule
×            & ×          &  513  &     63.85       \\
\checkmark   & ×          &   8 &     60.99        \\
×            & \checkmark &   4 &     58.36       \\
\checkmark   & \checkmark &   12 &     \textbf{65.02}       \\
\checkmark   & \checkmark &   24 &     61.92        \\
\bottomrule
\end{tabularx}
\end{table}

\begin{table}[h]
\caption{Number of Global Queries and Detail Queries}
\label{tab:query}
\begin{tabularx}{\linewidth}{CCCCc}
\toprule
$n_g$  & $n_d$ & $K$ & Total & Avg. Acc.  \\
\midrule
4   & 2 &  48  &     6     &   62.68  \\
8   & 4 &  96  &     12     &   \textbf{65.02}   \\
8   & 8 &  144 &     16     &   62.38  \\
16  & 8 &  144  &     24     &   61.92  \\
\bottomrule
\end{tabularx}
\end{table}
\subsection{Ablation Study}
In this section, we conduct ablation experiments on the generative 3D object classification task and report the average accuracy.
\paragraph{Impact of GSC and ADM}
Tab.~\ref{tab:hcc} shows that GSC reduces tokens from 513 to 8, while ADM contributes 4 tokens. Their combination achieves optimal performance with 12 tokens, demonstrating complementary roles in balancing compression and detail preservation. Increasing to 24 tokens degrades performance to 61.92\%, indicating that excessive tokens introduce redundancy and impair discriminative feature learning.

\paragraph{Query Configuration}
Tab.~\ref{tab:query} reveals that 8 global queries ($n_g=8$) and 4 detail queries ($n_d=4$) yield the best accuracy (65.02\%). Insufficient queries ($n_g=4$, $n_d=2$) cause 2.34\% performance drop, while excessive queries provide no benefit. Equal distribution ($n_g=8$, $n_d=8$) with 16 total tokens achieves only 62.38\%, highlighting the importance of balanced query allocation.

\begin{table}[h]
\caption{Feature Selection Strategy in ADM}
\label{tab:adm}
\begin{tabularx}{\linewidth}{ccCc}
\toprule
\makecell{Selection\\Method}  & \makecell{Token\\count} & Avg. Acc. & \makecell{Training Time\\(hours)} \\
\midrule
Select all  & 96 & 62.85 &  24.2   \\
Random   & 24 & 61.42 &  15.4   \\
Attention-only   & 4 & 63.42 &  11.8   \\
MLP-only  & 4 & 62.83 &  11.9   \\
ADM (Full)   & 4 & \textbf{65.02} &  \textbf{11.9}  \\
\bottomrule
\end{tabularx}
\end{table}
\paragraph{Feature Selection in ADM}
Tab.~\ref{tab:adm} compares selection strategies. Our ADM combines attention and MLP signals, achieving 65.02\% accuracy with only 4 tokens, outperforming random or single-signal methods. This dual-signal approach captures complementary features missed during global compression while identifying semantically rich regions, proving essential for maintaining performance under extreme compression.

\section{Conclusion}
We present HCC-3D, a hierarchical compensation compression framework that addresses computational bottlenecks in 3D vision-language models through global structure preservation and adaptive detail extraction. Our method achieves 98\% compression of 3D visual tokens (from 513 to 12) while attaining state-of-the-art performance across multiple benchmarks. The complementary Global Structure Compression (GSC) and Adaptive Detail Mining (ADM) modules preserve critical geometric information during aggressive compression, reducing training time by 52\% while improving accuracy. This work demonstrates that careful architectural design can overcome the efficiency-performance trade-off in 3D vision-language models, enabling deployment in resource-constrained scenarios and advancing scalable multimodal AI systems.

\section{Acknowledgments}
This work was supported in part by the Shandong Natural Science Foundation (Grant No. ZR2023QF046, No. ZR2023MF008), the National Natural Science Foundation of China (Grant No. 62301613, No. 62372468), the Taishan Scholar Program of Shandong (Grant No. tsqn202306130), and the Major Basic Research Projects in Shandong Province (Grant No. ZR2023ZD32).

\bibliography{aaai2026}

\end{document}